\documentclass[journal]{IEEEtran}

\ifCLASSINFOpdf
\else
\fi

\usepackage{graphicx}
\usepackage{amsmath,amssymb,amsfonts}
\usepackage{multirow}
\usepackage{hyperref}
\usepackage{epstopdf}
\usepackage{color}

\begin{document}
\title{A Prior Guided Adversarial Representation Learning and Hypergraph Perceptual Network for Predicting Abnormal Connections of Alzheimer's Disease}
\author{Qiankun Zuo, Baiying Lei, Shuqiang Wang, Yong Liu, Bingchuan Wang, Yanyan Shen
%\thanks{Corresponding author: Shuqiang Wang (email: sq.wang@siat.ac.cn)}
\thanks{Qiankun Zuo, Yanyan Shen and Shuqiang Wang are with the Shenzhen Institutes of Advanced Technology, Chinese Academy of Sciences, Shenzhen, 518055, China}
\thanks{Baiying Lei is with the School of Biomedical Engineering, Shenzhen University, Shenzhen, 518060, China}
\thanks{Yong Liu is with the Gaoling School of Artificial Intelligence, Renmin University of China, Beijing, 100872, China}
\thanks{Bingchuan Wang is with the School of Automation, Central South University, Changsha, 410083, China}}

\markboth{IEEE TRANSACTIONS ON NEURAL NETWORKS AND LEARNING SYSTEMS}%
{Shell \MakeLowercase{\textit{et al.}}: Bare Demo of IEEEtran.cls for IEEE Journals}

\maketitle

\begin{abstract}
Alzheimer's disease is characterized by alterations of the brain's structural and functional connectivity during its progressive degenerative processes. Existing auxiliary diagnostic methods have accomplished the classification task, but few of them can accurately evaluate the changing characteristics of brain connectivity. In this work, a prior guided adversarial representation learning and hypergraph perceptual network (PGARL-HPN) is proposed to predict abnormal brain connections using triple-modality medical images. Concretely, a prior distribution from the anatomical knowledge is estimated to guide multimodal representation learning using an adversarial strategy. Also, the pairwise collaborative discriminator structure is further utilized to narrow the difference of representation distribution. Moreover, the hypergraph perceptual network is developed to effectively fuse the learned representations while establishing high-order relations within and between multimodal images. Experimental results demonstrate that the proposed model outperforms other related methods in analyzing and predicting Alzheimer's disease progression. More importantly, the identified abnormal connections are partly consistent with the previous neuroscience discoveries. The proposed model can evaluate characteristics of abnormal brain connections at different stages of Alzheimer's disease, which is helpful for cognitive disease study and early treatment.
\end{abstract}


% \fi
\IEEEpeerreviewmaketitle

\section{Introduction}
\IEEEPARstart{A}{lzheimer's disease} (AD) is one of the most common neurodegenerative diseases among the elderly~\cite{ref_01}. It is reported that the number of people living with AD will drastically rise from 50 million in 2018 to 152 million in 2050 around the world~\cite{ref_02}. AD has become a major social problem that endangers public health and hinders economic development in the coming years~\cite{ref_02_1}. However, the exact cause of that disease is unclear, and no treatments or effective drugs are reported to cure the AD~\cite{ref_03,ref_031,ref_032}. Therefore, much attention is paid to the early diagnosis of AD~\cite{ref_03_0, ref_03_00, ref_03_1, ref_03_2, ref_03_3, ref_03_4}, so that timely intervention can be taken to slow down the progression of AD.
%~\cite{ref_04}

Neuroscience studies~\cite{ref_04_1, ref_04_2, ref_04_3} on the brain have shown characteristic changes in brain morphology, structural or functional connections at the early stage of AD. Brain network is suitable to describe these characteristics by employing different imaging modalities, such as T1-weighted magnetic resonance imaging (MRI), positron emission tomography (PET), resting-state functional magnetic resonance imaging (fMRI), and diffusion tensor imaging (DTI). Moreover, it can be appropriately characterized by graph theory, where the nodes represent the spatially distributed regions of interest (ROIs) and the edges represent the link relationship between ROIs or subjects~\cite{ref_05}. Previous works based on graph convolutional network (GCN)~\cite{ref_07} extracted features for each ROI or subject and then built a classifier for AD diagnosis by using either unimodal or bimodal imaging data. They can achieve good classification performance and analyze the pathological brain regions associated with cognitive disease. However, those works neglect to unravel the underlying interactions among multiple disease-related ROIs in the brain. Besides, in order to analyze cognitive disease, hypergraph~\cite{ref_08} based methods were used to improve the classification performance and produce discriminative connections by considering high-order relations among multiple ROIs from unimodal imaging or among multiple subjects from multimodal imaging. The extracted features by these methods neglect the high-order relations of multiple ROIs both within and between multimodal images, which is essential for cognitive disease analysis.

Recently, multimodal image fusion has attracted much attention in disease diagnosis, because it can provide complementary information about the disease and thus improve disease detection performance~\cite{ref_09,ref_09_1,ref_09_2, ref_09_21}. Besides, machine learning is widely used in medical image processing~\cite{ref_09_22,ref_09_23,ref_09_24, ref_09_25}. Generative Adversarial Network (GAN) based on variational inference~\cite{ref_09_3,ref_09_31,ref_09_4,ref_09_5} is strongly capable of generalization and distribution fitting ability in image analysis~\cite{ref_09_6,ref_09_7,ref_09_8} and graph representation learning~\cite{ref_10}, which has been enhanced in terms of robustness by incorporating normal distribution~\cite{ref_11}. Considering that the neurological activities associated with cognition and behavior are a result of the joint mutual interaction among multiple brain regions~\cite{ref_12,ref_13}, hypergraph is more appropriate to characterize that high-order relations among multiple ROIs. Inspired by the above observations, in this paper, we propose a novel prior guided adversarial representation learning and hypergraph perceptual network (PGARL-HPN) for predicting abnormal brain connections of different cognitive disease stages. The proposed model can automatically learn robust representations and produce discriminative united connectivity representing the overall brain network of multimodal images. Based on anatomical knowledge, specific disease-related ROIs are applied to the graph data to estimate prior distribution. A bidirectional adversarial mechanism is introduced to stabilize the representation learning and speed up the convergence for multimodal representation learning by incorporating the estimated prior distribution. The pair samples from data space and representation space are sent to the designed pairwise collaborative discriminator to preserve the sample consistency and distribution consistency. Also, the reconstruction and classification modules are utilized to make representations class-discriminative and robust. Besides, a hypergraph-based network is employed to generate hyperedges for each imaging modality and then produce fused representation by perceptual convolution, which captures the high-order relations among multiple ROIs within single modality imaging and bridges the high-order relations between multimodal images. As a result, the united connectivity-based features are extracted to retain disease-related complementary information and improve abnormal connection prediction performance. The main contributions of this framework are as follows:

\begin{itemize}
	\item A Priori Guided Adversarial Representation Learning (PGARL) module is designed to learn latent representations from multimodal images. It can use the estimated prior distribution to guide the bidirectional adversarial network in an optimal manner, thus stabilizing the representation learning and speeding up the convergence.
	\item The Pairwise Collaborative Discriminator (PCD) is introduced to associate the edge and joint distribution between the input data and latent representations, which can minimize the difference of representation distribution from multimodal images.
	\item Hypergraph Perceptual Network (HPN) is developed to establish high-order relations between and within multimodal images, improving the fusion effects in terms of morphology-structure-function information and enhancing the discrimination of united connectivity-based features.
\end{itemize}

The remaining parts of this paper are organized as follows. Section II presents the related works. Section III describes the details of the proposed method. Section IV first introduces experimental settings and competing methods, and then presents results on the public database. Section V discusses the reliability of our results and the limitations of current study. Section VI concludes this paper and future work.

\section{Related Work}
The current research of AD diagnosis using the GCN approach can be divided into two categories: group-based approach and individual-based approach. The first approach constructs one graph for all the subjects by treating every subject as one node. For instance, Parisot et al.~\cite{ref_29} extracted node features by applying convolutional neural networks (CNN) on MRI of each subject and built node connections using non-image data (e.g., sex, age), then the constructed graph is utilized to refine node features for AD diagnosis through semi-supervised learning. In order to make use of complementary information from multimodal images, Yu et al.~\cite{ref_30} presented a multi-scale enhanced GCN combining with the fMRI, DTI, and non-image data for AD study. Another work in~\cite{ref_31} established a novel framework that adds extra-label information to build a graph, which effectively improves the performance of Alzheimer's disease prediction. In the second approach, each subject is built as one graph by predefined ROIs based on a specific atlas in the brain. Yu et al.~\cite{ref_32} constructed a brain network model with weighted graph regularized sparse constraints, which yields visually enhanced accuracy in Mild Cognitive Impairment (MCI) classification. Also, the addition of more modal images can help disease diagnosis accuracy. Lei et al.~\cite{ref_33} built a self-calibrated brain network combining with fMRI and DTI to learn functional and structural complementary features. Xing et al.~\cite{ref_34} trained a GCN model to study the tissue-functional complementary characteristics of brain networks by using MRI and fMRI. However, the existing data-driven methods have accomplished the task of sample classification and prediction, which may ignore accurate evaluation on the varied characteristics of brain network and lack of biological explanations.

In the field of cognitive disease analysis, the hypergraph-based methods can be separated into two groups: unimodal approach and multimodal approach. Researchers explore high-order relations among brain regions by constructing a hypergraph using single modality imaging in the first approach. For example, Jie et al.~\cite{ref_35} utilized fMRI to generate hyper-connectivity features, which significantly improve the performance of MCI diagnosis and further discover valuable biomarkers. The work in~\cite{ref_36} improved the construction of hypergraph from fMRI by applying sparse constraints and hyperedge weighting, and it provides additional information about the underlying biomarkers for cognitive study. In the second approach, multimodal images are jointly used to construct multiple hypergraphs for modeling high-order relationships among subjects. In order to overcome the problem of modal incompleteness, Liu et al.~\cite{ref_37} presented a hypergraph-based method to bridge relationships among subjects from MRI, PET, and cerebrospinal fluid (CSF) for automatic brain disease diagnosis. Zhu et al.~\cite{ref_38} adopted an iterative scheme in hypergraph learning using MRI and PET to identify the diagnostic labels and predict clinical scores of degenerative disease. Li et al.~\cite{ref_39} built multimodal hyper-networks from two modalities of fMRI, and it achieves good performance and generates discriminative connections between MCI and normal control (NC). Nevertheless, the primary deficiency of the above methods is that they may not consider the high-order relations both within and between multimodal images, which do not entirely explore the potential complementary information from data.

\section{Method}

\begin{figure*}[htbp]
	\centering
	\includegraphics[scale=0.62]{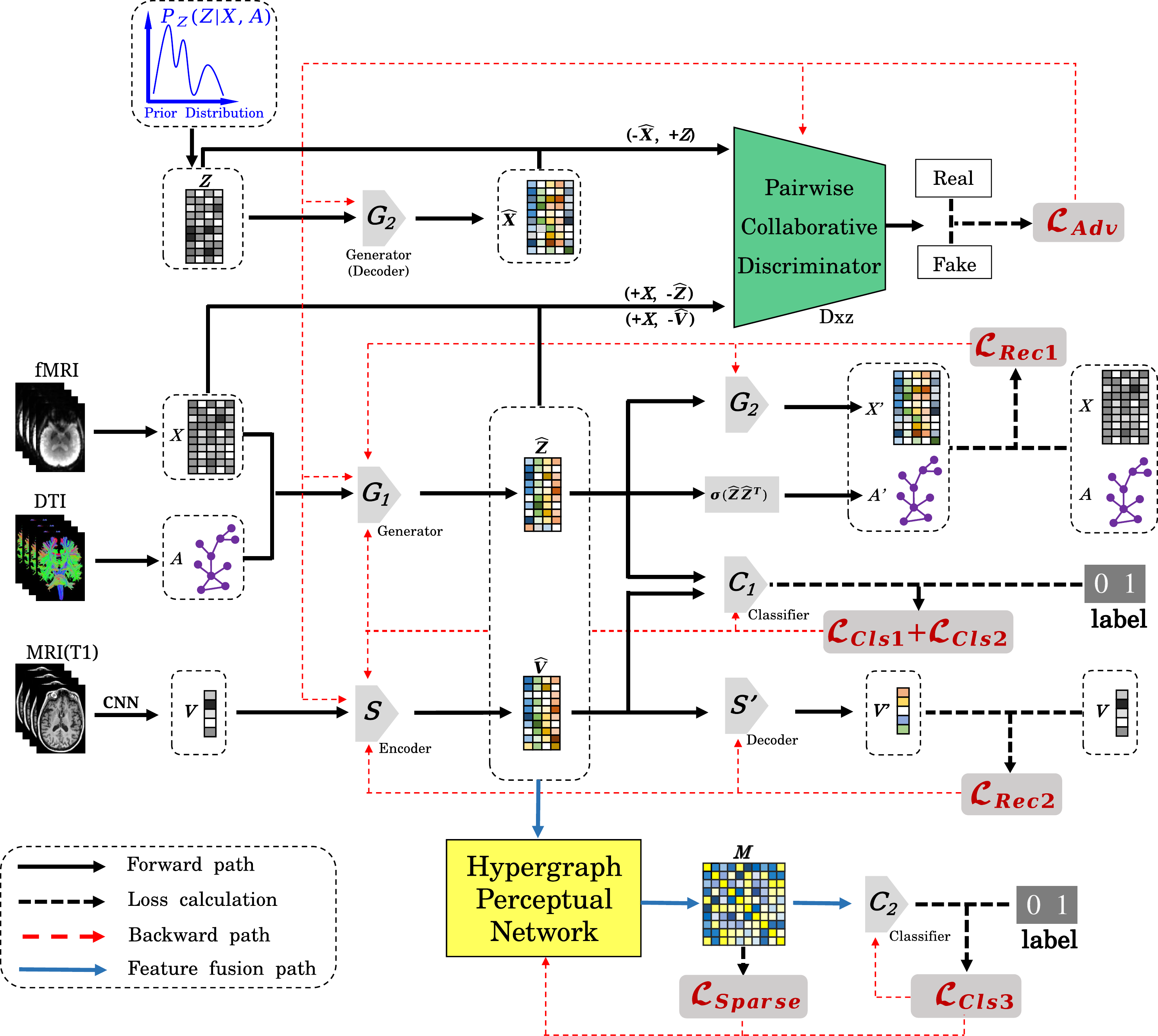}
	\caption{Overall framework of the proposed PGARL-HPN for AD diagnosis using fMRI, DTI, and MRI. $A$ and $A'$ represent the structural connectivity (SC) and reconstructed SC matrix, respectively. $X$ and $X'$ denote the functional time series (FTS) and the reconstructed FTS feature, respectively. $V$ and $V'$ are the feature vector (FV) and reconstructed FV. $P_Z(Z|X, A)$ denotes the estimated prior distribution, "+" and "-" mean the real and fake samples. $\widehat{Z}$ and $\widehat{V}$ represent latent representations. $M$ is united connectivity that represents the overall brain network of multimodal images.}
	\label{fig1}
\end{figure*}

%\begin{figure}[!t]
%\centering
%\includegraphics[width=2.5in]{myfigure}
% where an .eps filename suffix will be assumed under latex,
% and a .pdf suffix will be assumed for pdflatex; or what has been declared
% via \DeclareGraphicsExtensions.
%\caption{Simulation results for the network.}
%\label{fig_sim}
%\end{figure}

\subsection{Overview}
Assuming subjects have the three modal images (i.e., fMRI, DTI, and MRI), our goal is to learn a complex non-linear mapping network to fuse multimodal images for predicting abnormal brain connections at different stages of Alzheimer's disease. As illustrated in Fig.~\ref{fig1}, our method consists of three components: 1) prior distribution estimation, 2) adversarial representation learning, and 3) the hypergraph perceptual network. Firstly, a kernel density estimation (KDE) method is introduced to estimate the prior distribution from graph data in terms of subject labels. Secondly, the estimated distribution is incorporated into a bidirectional adversarial learning network for multimodal representation learning. It should be stressed that the designed pairwise collaborative discriminator is used to constrain the representations in a joint embedding space. Then, a hypergraph-based network is developed to fuse the learned representations to produce united connectivity-based features. Finally, our network is trained with the following objective functions: the adversarial loss, the reconstruction loss, the classification loss, and the sparse regularization loss. The goal of loss functions is to improve the representation learning ability and the effect of multimodal image fusion. Details of the architecture and the hybrid objective functions are described in the following sections: ~\ref{sec: architectures} and ~\ref{sec: objective}.

\subsection{Architectures}
\label{sec: architectures}
\subsubsection{Prior Distribution Estimation}
Suppose an indirect graph $\mathcal{G}(\mathcal{V},\mathcal{E})$ is formed with $N$ brain ROIs, where $\mathcal{V}=\{\nu_1,\nu_2,...,\nu_N\}$ and $\mathcal{E}=\{\varepsilon_1,\varepsilon_2,...,\varepsilon_N\}$ are a set of nodes and edges respectively. Specifically, $X=\{x_1,x_2,...,x_N\}\in\mathbb{R}^{N \times d}$ denotes the node feature derived from fMRI, and $A\in\mathbb{R}^{N \times N}$ represents the edge connectivity computed from DTI. The element in adjacent matrix $A$ is defined as $A_{ij}=1$ if there exists connection between $i$th and $j$th region, otherwise $A_{ij}=0$.

Normal distribution $\mathbb{N}(0,1)$ cannot represent the distribution of graph data properly, and an appropriate distribution $P_Z(Z)$ can help the adversarial network to boost representation learning ability. There is no other prior information except for the given graph data (i.e., $X$, $A$). It is possible to estimate a prior distribution from the graph data. The heterogeneity between $A$ and $X$ makes it difficult to obtain $P_Z(Z|X, A)$ directly, but they are structurally consistent. A cross-domain prototypes method can eliminate the gap between the node domain (i.e., $X$) and edge domain (i.e., $A$). Thus $P_Z(Z|X,A)$ can be replaced with $P_Z(Z|X_{U_0},A_{U_0})$ , where $U_0$ means the index set of prototypes.

The Determinantal Point Process(DPP)~\cite{ref_42} based prototype learning method is adopted to select a diversified prototype subset $U_0$. Specifically, based on some certain ROIs that have been verified to be closely related to the disease, a set of nodes $U\subseteq\mathcal{V}$ is chosen from the connectivity matrix $A$.  When given the subset size $m$, the sampling probability of $U$ is written as:
\begin{equation}
	P_A(U) = \frac{det(A_U)}{\sum_{|U'|=m} det(A_{U'})}
\end{equation}
Where, $A_U\equiv[A_{ij}]_{i,j\in\mathnormal{U}}$, $det(\cdot)$ denotes the determinant of a square matrix. Hence, the maximum value of $P$ can output a subset of nodes that reflect the main features of the graph.

According to the prototype index set $U_0$ with $|U_0|=m$, the corresponding node features are sampled from $X$ to form features matrix $X_{U_0}\in\mathbb{R}^{m \times d}$. Then the feature dimension is reduced by Principal Component Analysis (PCA) to get $Z_{U_0}\in\mathbb{R}^{m \times q}$. $q$ is the dimension in latent representation space. A non-parametric estimation method (i.e., KDE) is introduced to estimate prior distribution. Assuming $z_i \in Z_{U_0}$ is a latent representation of each node, $P_Z(Z|X_{U_0},A_{U_0})$ is defined as
\begin{equation}
	\begin{split}
		P_Z(Z|X_{U_0},A_{U_0}) &= \frac{1}{m} \sum_{i=1}^{m} K_b({Z-z_i}) \\
		&= \frac{1}{mb} \sum_{i=1}^{m} K(\frac{Z-z_i}{b})
	\end{split}
\end{equation}
Where $K(\cdot)$ is a multi-dimensional Gaussian kernel function, $b$ denotes the bandwidth that determines the smoothness of the estimated prior distribution.

At last, the approximation of prior distribution $P_Z(Z)$ is obtained by the following formula
\begin{equation}
	Pz(Z) \rightarrow Pz(Z|X,A) \rightarrow Pz(Z|X_{U_0},A_{U_0})
\end{equation}

\begin{figure*}[htbp]
	\centering
	\includegraphics[scale=0.6]{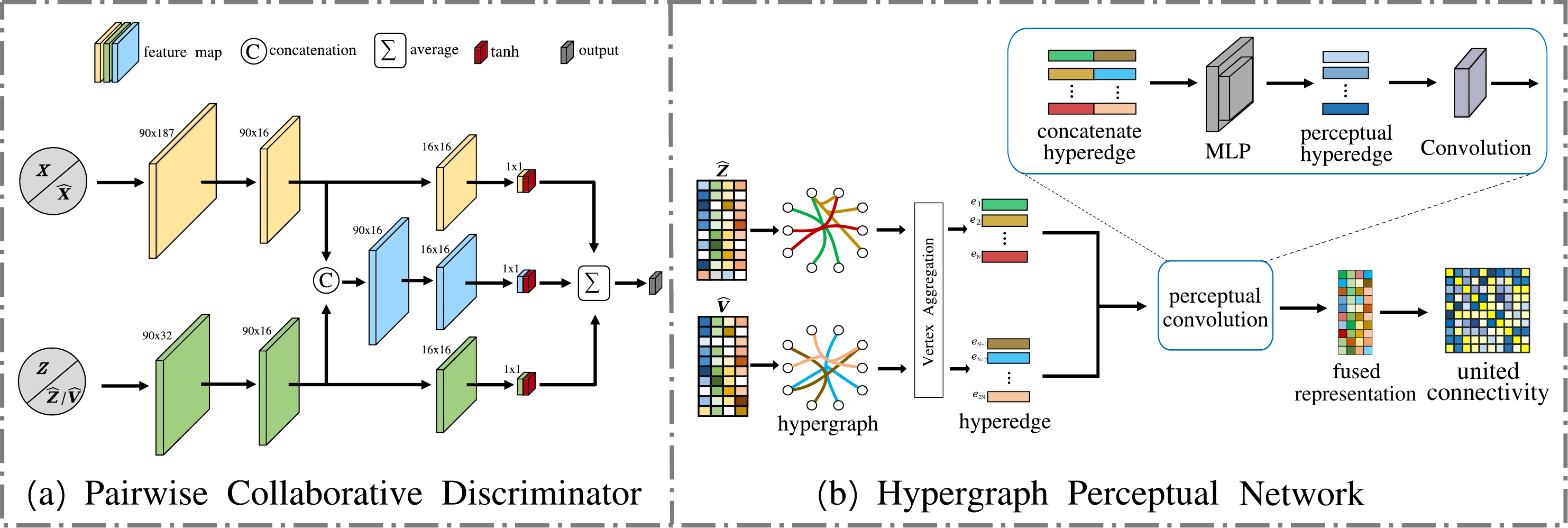}
	\caption{The detailed structure of (a) Pairwise Collaborative Discriminator and (b) Hypergraph Perceptual Network.}
	\label{fig2}
\end{figure*}

\subsubsection{Adversarial Representation Learning}
The multimodal data is limited, so it is challenging to learn discriminative latent representations for disease prediction.  In order to solve this problem, a bidirectional adversarial mechanism is introduced to incorporate the estimated distribution to augment samples, which can enhance the ability of multimodal representation learning. Specifically, we design a pairwise collaborative discriminator for learning latent representations and keeping their unimodal features by applying reconstruction and classification modules. This part has three phases: the adversarial phase, the reconstruction phase, and the classification phase.

In the adversarial phase, the generator $G_1$ accepts $A$ and $X$ as inputs and outputs a latent representation matrix $\hat{Z}\in\mathbb{R}^{N \times q}$. Meanwhile, the encoder $S$ accepts $A$ and $V\in\mathbb{R}^{1 \times q}$ as inputs and outputs a latent representation $\hat{V}\in\mathbb{R}^{N \times q}$. The data $Z\in\mathbb{R}^{N \times q}$ is sampled from the prior distribution $P_Z(Z|X,A)$, which is sent to the generator $G_2$ to get $\hat{X}\in\mathbb{R}^{N \times d}$. After that, three pairs of data, ($\hat{X}$,$Z$), ($X$, $\hat{Z}$) and ($X$,$\hat{V}$), are input into the discriminator $D_{XZ}$ for adversarial training. $X$ and $Z$ are positive samples. $\hat{X}$, $\hat{Z}$ and $\hat{V}$ are negative samples. In particular, GCN with two-layer is used as the generator $G_1$, $G_2$, and the encoder $S$.

The structure of pairwise collaborative discriminator $D_{XZ}$ is shown in Fig.~\ref{fig2}(a). The developed discriminator consists of two separate subnetworks (upper and lower) and a joint subnetwork (middle). All the subnetworks have three layers. From the first layer to the third layer, the filter size of the upper subnetwork is $187\times1, 1\times90, 16\times16$; the filter size of the lower subnetwork is $32\times1, 1\times90, 16\times16$; the filter size of the middle subnetwork are $32\times1, 1\times90, 16\times16$. The number of filters is 16, 16, and 1 for each layer of the subnetworks. In the last layer of each subnetwork, the $tanh$ activation function is used to constrain output value in the range of $-1\sim1$. The separative subnetworks can supervise the generators to learn robust representations, while the added joint subnetwork significantly improves the optimization efficiency.

The reconstruction phase is used to stabilize the representation learning. $\hat{Z}$ is fed to $G_2$ to rebuilt $X'\in\mathbb{R}^{N \times d}$, and reconstruct $A'\in\mathbb{R}^{N \times N}$ through matrix inner product operation $\sigma(\hat{Z}\hat{Z}^T)$. $\hat{V}$ is fed to the decoder $S'$ to construct $V'\in\mathbb{R}^{1 \times q}$. Here, the $G_2$ and the decoder $S'$ are set two-layer GCN with different parameters in hidden layer.

In the classification phase, to obtain more class-discriminative features, a classifier $C_1$ is designed to constrain the latent representations $\hat{Z}$ and $\hat{V}$. The network of $C_1$ is defined as the same in ~\cite{ref_421}, where the representation (i.e., $\hat{Z}$ or $\hat{V}$) is first averaged along the dimension direction and then sent to a two-layer Multi-Layer Perception (MLP) for classification.

\subsubsection{Hypergraph Perceptual Network}
Compared with the conventional graph characterized by pairwise relationships between paired nodes, hypergraph is more appropriate for capturing high-order relations among multiple brain regions. Hence, a hypergraph-based network combined with MLP and convolution is utilized to fuse the learned representations (i.e., $\hat{Z}$ and $\hat{V}$). The obtained united connectivity matrix is sent to the classifier $C_2$ for task learning. The detail of the network is illustrated in Fig.~\ref{fig2}(b).

First, a hypergraph is constructed for each learned representation. By defining a hyperedge as one centered node connecting multiple other nodes, we can construct two hypergraphs with a total of $2N$ hyperedges. Specifically, we use the K-nearest neighbor(KNN) method to select the nodes for each hyperedge based on the Euclidean distance. At last, we can get an incident matrix $H_1$ and $H_2$ from the representations $\hat{Z}$ and $\hat{V}$, respectively.  Assuming hyperedge set $E=\{e_1,e_2,...,e_N\}\in\mathbb{R}^{N \times N}$, the incident matrix can be represented by
\begin{equation}
	H(\mathcal{V},E)=\{_{1,if\   \nu \notin E}^{0, if\   \nu \in E}
\end{equation}

Then, we obtain the hyperedge features by vertex aggregation. The hypergraph convolution is spitted into convex convolution and hyperedge convolution. The former is a non-parameter operation, which is illustrated with the following formula
\begin{equation}
	\hat{Z}_E=D_{1e}^{-1/2} H_1^T D_{1e}^{-1/2} \hat{Z}
\end{equation}
\begin{equation}
	\hat{V}_E=D_{2e}^{-1/2} H_2^T D_{2e}^{-1/2} \hat{V}
\end{equation}
Where, $D_{1e}$ and $D_{2e}$ are the edge degree of $H_1$ and $H_2$, respectively.

Next, we design a novel network that combines multi-layer perception and graph convolution to capture the high-order relations between and within multimodal images. The concatenate hyperedge is fused between multimodal images by the MLP network, and the graph convolution network transforms the fused hyperedge into vertex features. The used MLP is a one-layer network, and the used GCN is a one-layer network. This operation process can be described as follows:
\begin{equation}
	F=D_{v}^{-1/2} H D_{v}^{-1/2} [(\hat{Z}_E || \hat{V}_E) \cdot W]
\end{equation}
Where, $H=(H_1 + H_2)/2$, $D_v$ is the node degree of $H$. $||$ means concatenating two feature matrices. $W$ is the MLP layer parameters.

Then, the fused features are used to construct brain united connectivity with the following form:
\begin{equation}
	M=\sigma(FF^T)
\end{equation}
Finally, the obtained united connectivity is flattened and sent to a classifier $C_2$ for the classification task. $C_2$ is a two-layer MLP network.

\subsection{Hybrid loss function}
\label{sec: objective}
Success in effectively fusing multimodal images requires joint representation learning. The outputs of generator $G_1$ and encoder $S$ should have the same distribution despite the heterogeneity of input data. Besides, the learned representations and fused representation must be discriminative concerning class labels, and the output united connectivity matrix of HPN is believed to be sparse. An extensive object function is designed to optimize the network to achieve this goal in the proposed model. There are four types of loss function: the adversarial loss, the reconstruction loss, the classification loss, and the sparse regularization loss. The detailed information will be described below.

The adversarial loss is used to keep the distribution-level consistency between the input data distribution and the latent representations. Given the node features $X \sim P_{fMRI}(X)$, the node adjacent matrix $A \sim P_{DTI}(A)$, the semantic features $V \sim P_{MRI}(V)$ and the estimated prior distribution $Z \sim Pz(Z|X,A)$, the loss objective can be expressed as follow

\begin{equation}
	\begin{split}
		\pounds_  {D} = &-\mathbb{E}_{{Z} \sim P_Z(Z|X,A)}[D_{XZ}(Z)] \\
		&-2\mathbb{E}_{{X} \sim P_{fMRI}(X)}[D_{XZ}(X)] \\
		&+ \mathbb{E}_{{Z} \sim Pz(Z|X,A)}[D_{XZ}(G_2(Z))] \\
		&+ \mathbb{E}_{A \sim P_{DTI}(A), X \sim P_{fMRI}(X)}[D_{XZ}(G_1(A,X))] \\
		&+ \mathbb{E}_{A \sim P_{DTI}(A), V \sim P_{MRI}(V)}[D_{XZ}(S(A,V))]
	\end{split}
\end{equation}

\begin{equation}
	\begin{split}
		\pounds_  {G} = &- \mathbb{E}_{{Z} \sim P_Z(Z|X,A)}[D_{XZ}(G_2(Z))] \\
		&- \mathbb{E}_{A \sim P_{DTI}(A), X \sim P_{fMRI}(X)}[D_{XZ}(G_1(A,X))] \\
		&- \mathbb{E}_{A \sim P_{DTI}(A), V \sim P_{MRI}(V)}[D_{XZ}(S(A,V))]
	\end{split}
\end{equation}

\begin{equation}
	\pounds_  {Adv} = \pounds_  {G} + 0.1 \pounds_  {D}
\end{equation}

Reconstruction loss functions are added to the adversarial learning network to retain the unimodal imaging information in learned representations. The reconstruction loss can be defined as follows:

\begin{equation}
	\begin{split}
		\pounds_  {Rec} &= \pounds_  {Rec1} + \pounds_  {Rec2} \\
		&=\mathbb{E}_{X \sim P_{fMRI}(X)}[f(X,X')] + \mathbb{E}_{A \sim P_{DTI}(A)}[f(A,A')] \\
		&\ \  \  +\mathbb{E}_{V \sim P_{MRI}(V)}[f(V,V')]
	\end{split}
\end{equation}
Where, $X'=G_2(G_1(X))$ and $A'=\sigma(G_1(X) G_1(X)^T)$, $V'=S'(S(V))$ are the reconstructed data, $f(x_1,x_2)=x_1 \cdot logx_2 + (1-x_1) \cdot log(1-x_2)$ is binary cross entropy function.

In order to ensure the learned representations be discriminative to the corresponding labels, the classification loss is defined by cross-entropy operation:

\begin{equation}
	\begin{split}
		&\pounds_  {Cls} = \pounds_  {Cls1} + \pounds_  {Cls2} + \pounds_  {Cls3}\\
		&=\mathbb{E}_{A \sim P_{DTI}(A), X \sim P_{fMRI}(X)}[y \cdot log(C_1(A,X))] \\
		&\ \ \ + \mathbb{E}_{A \sim P_{DTI}(A), V \sim P_{MRI}(V)}[y \cdot log(C_1(A,V))] \\
		&+ \mathbb{E}_{A \sim P_{DTI}(A),X \sim P_{fMRI}(X),V \sim P_{MRI}(V)}[y \cdot log(C_2(A,X,V))]
	\end{split}
\end{equation}
Where, $y$ is the true label.

In addition, the $L1$ penalty is introduced to make the united connectivity matrix sparse. It is given below:

\begin{equation}
	\begin{split}
		&\pounds_  {Sparse} = || M ||_1
	\end{split}
\end{equation}
Here, $|| \cdot ||_1$ is the $L1$ norm.

In Conclusion, the total loss of the proposed frame is:
\begin{equation}
	\pounds_{total} = \pounds_{Adv} + \pounds_{Rec} + \pounds_{Cls} + \lambda \pounds_  {Sparse}
\end{equation}
Where, $\lambda$ is a hyper-parameter that determines the relative importance of sparse loss items.

\section{Experiments}
\label{sec:Experiments1}

\subsection{Data description and preprocessing}

In this study, a total of 300 subjects are collected from Alzheimer ’s Disease Neuroimaging Initiative (ADNI-3)\footnote{http://adni.loni.usc.edu/} database,including early mild cognitive impairment(EMCI), late mild cognitive impairment (LMCI), Alzheimer Disease(AD), and normal control (NC). The collected subjects with complete three modal images (i.e., fMRI, DTI, T1-weighted MRI) are scanned by a 3T MRI scanner at different sites. For fMRI data, the range of image resolution in X and Y dimensions ranges from 2.5mm to 3.75mm, the range of slice thickness is from 2.5mm to 3.4mm. The TR(time of repetition) ranges from 0.607s to 3.0s, and the TE (time of echo) value is in the range of 30ms to 32ms. The total length of scan time is 10 minutes. For DTI data, the range of image resolution in X and Y dimensions ranges from 0.9mm to 2.7mm, the slice thickness is 2.0mm. The TR value is between 3.4s and 17.5s, and the TE value is in the range of 56ms to 105ms. The gradient directions of DTI data are between 6 and 126. For T1-weighted MRI, the range of image resolution in X and Y dimensions ranges from 1.0mm to 1.06mm, the range of slice thickness is from 1.0mm to 1.2mm. The TR value is 2.3s, and the TE value is in the range of 2.94ms to 2.98ms. The detailed information is shown in Table~\ref{Table0}.

\begin{table}[htbp]
	\caption{Detailed information about the subjects in this study.\newline SD: Standard Deviation}\label{Table0}
	\renewcommand\arraystretch{1.5}
	\begin{center}
		\begin{tabular}{ccccc}
			\hline \cline{1-5}
			Group & NC(78) & EMCI(82) & LMCI(76) & AD(64)\\
			\hline
			Male/Female & 39M/39F & 40M/42F & 43M/33F & 39M/25F\\
			Age(mean$\pm$SD) & 76.0/8.0 & 75.9/7.5 & 75.8/6.4 & 74.7/7.6\\
			\hline
		\end{tabular}
	\end{center}
\end{table}

For fMRI data preprocessing, we adopt the standard procedures using the GRETNA toolbox~\cite{ref_43} to filter the functional time-series signal. The main steps include magnetization equilibrium calibration, head-motion artifacts correction, spatial normalization, a band-pass filtering between 0.01Hz and 0.08Hz. Based on the Automated Anatomical Labeling(AAL) atlas~\cite{ref_44}, a total of 90 non-overlapping ROIs were mapped to segment the brain. After that, we normalize the time-series signal into a same length and obtain one matrix for each subject with the dimension size $90\times187$, which is the input feature matrix $X$ of our model.

For DTI structural brain network, PANDA toolbox~\cite{ref_45} is used to perform the prepossessing operation for the determination of brain fractional anisotropy. There are five main steps with default settings, including resampling, skull removal, gap cropping, head movements and eddy currents correction. By setting the tracking conditions, network nodes and tracking stopping conditions, it generates a structural network matrix based on the deterministic fiber tracking method. The obtained matrix with the dimension size $90\times90$ is the input adjacent matrix $A$ of our model.

The downloaded T1-weighted MRI images are pre-processed with the following proprocessing steps, including Brain Extraction Tool(FSL-BET)~\cite{ref_46_1} that strip non-brain tissue of the whole head, and FSL-FLIRT~\cite{ref_46_2} that align the images to the standardized template. The output image with a voxel size of $91\times109\times91$ in Neuroimaging Informatics Technology Initiative(NIFTI) file format is then input into a designed 40-layer DenseNet~\cite{ref_47} model to extract semantic vector. The obtained vector feature with the dimension size $1\times32$ is the input feature $V$ of our model.

\begin{table*}[htbp]
	\centering
	\renewcommand\arraystretch{1.2}
	\caption{Mean prediction performance of the proposed and other related methods.(\%)}\label{TableII}
	\begin{tabular}{c|c|cccc|cccc|cccc}
		\hline
		\multirow{2}{*}{Modality}   & \multirow{2}{*}{Method} & \multicolumn{4}{c|}{AD vs. NC} & \multicolumn{4}{c|}{LMCI vs. NC} & \multicolumn{4}{c}{EMCI vs. NC} \\ \cline{3-14}
		&                         & Acc    & Sen   & Spec  & Auc   & Acc    & Sen    & Spec   & Auc   & Acc    & Sen    & Spec   & Auc   \\ \hline
		\multirow{2}{*}{fMRI}       & N2EN                    & 79.57  & 70.31 & 81.81 & 87.68 & 74.02  & 72.36  & 74.32  & 80.75 & 72.50  & 68.29  & 75.67  & 79.57 \\ \cline{2-14}
		& Ours                 & 80.98  & 76.56 & 80.33 & 88.92 & 74.68  & 76.32  & 73.41  & 82.79 & 73.12  & 78.04  & 71.91  & 81.31 \\ \hline
		\multirow{3}{*}{fMRI \&DTI} & MPCA                    & 80.28  & 70.31 & 88.64 & 85.37 & 75.97  & 73.68  & 78.20  & 80.58 & 74.37  & 79.27  & 69.23  & 83.22 \\ \cline{2-14}
		& DCNN                    & 84.51  & 87.50 & 82.05 & 89.40 & 79.87  & 77.63  & 82.05  & 84.29 & 76.25  & 76.83  & 75.64  & 82.68 \\ \cline{2-14}
		& Ours               & 88.73  & 84.37 & 92.31 & 97.48 & 84.42  & 84.21  & 84.62  & 94.01 & 82.50  & 82.93  & 82.05  & 91.65 \\ \hline
		\multirow{2}{*}{fMRI \& DTI \& MRI} & SPMRM & 94.37 & 95.31 & 93.59  & 97.02  & 90.91  & 96.05  & 85.90  & 89.05  & 83.75  & 90.24 & 76.92  & 87.74 \\ \cline{2-14}
		& {\bfseries Ours}  & {\bfseries 96.47}  & {\bfseries 98.43} & {\bfseries 94.87} & {\bfseries 99.59} & {\bfseries 92.20}  & {\bfseries 96.05}  & {\bfseries 88.46}  & {\bfseries 94.83} & {\bfseries 87.50}  &  {\bfseries 92.68}  & {\bfseries 82.05}  & {\bfseries 92.72} \\ \hline
	\end{tabular}
\end{table*}

\subsection{Experimental settings}
In this study, we use three kinds of binary classification tasks, i.e., (1) EMCI vs. NC, (2) LMCI vs. NC, (3) AD vs. NC. 10-fold cross-validation is selected for task learning. In order to demonstrate the superiority of our proposed model compared with other models, we introduce previous methods for comparison. (1) fMRI based methods, including non-negative elastic-net based method(N2EN)~\cite{ref_420} and Ours. (2) fMRI\&DTI based methods, including Multilinear Principal Component Analysis (MPCA)~\cite{ref_49}, Diffusion convolutional neural networks (DCNN)~\cite{ref_421} and Ours. (3) fMRI\&DTI\&MRI based methods: self-paced sample weighting based multi-modal rank minimization (SPMRM)~\cite{ref_49_0} and Ours. In the SPMRM method, we replace the PET and CSF data with structural connectivity and functional time series, respectively.

In the experiments, the proposed model is trained on the three modal images. we set the model parameter as follows: $N=90, d=187, q=32, m=10$. Four certain ROIs (i.e., the left and right hippocampus, the left and right Parahippocampal) are included in the estimation of the prior distribution. Tanh and sigmoid activation functions are used for generators and decoders, respectively. The batch size of the model is set at 8. In the training process, TensorFlow1\footnote{http://www.tensorflow.org/}  is utilized to implement on an NVIDIA TITAN RTX2080 GPU device. It takes about 8 hours to train our model on each fold cross-validation with a total of 500 epochs. The initial learning rate of the generators, encoder, decoder, and classifiers is $10^{-3}$ and will decrease to $10^{-4}$ at 100 epochs; while the learning rate of the discriminator is set 0.0001 constantly. The learning rate of the hypergraph perception network is set 0 at the beginning 100 epochs and then decreased by multiplying $(1-iter/max_{iter})^{power}$ along with the iteration with an initial value of 0.001. The momentum method with a coefficient of 0.9 was used to optimize the learning processes.

In the prediction performance evaluation, we use four metircs to quantitativly evaluate the diagnosis performance, including accuracy(ACC), sensitivity(SEN), specificity(SPE), and area under the receiver operating characteristic(ROC) curve. The Area Under a ROC curve (AUC) is used to comprehensively measure classifier performance ($0\le AUC\le 1$). Note that an AUC value of 0.5 indicates a random classifier.

%The metrics are defined as below:
%
%\begin{equation}
%ACC = (TP + TN) / (TP + FN + TN + FP)
%\end{equation}
%\begin{equation}
%SEN = TP / (TP + FN)
%\end{equation}
%\begin{equation}
%SPE = TN / (TN + FP)
%\end{equation}
%Where, $TP, TN, FP$, and $FN$ as true positive, true negative, false positive, and false negative respectively.

\begin{figure}[htbp]
	\centerline{\includegraphics[width=\columnwidth]{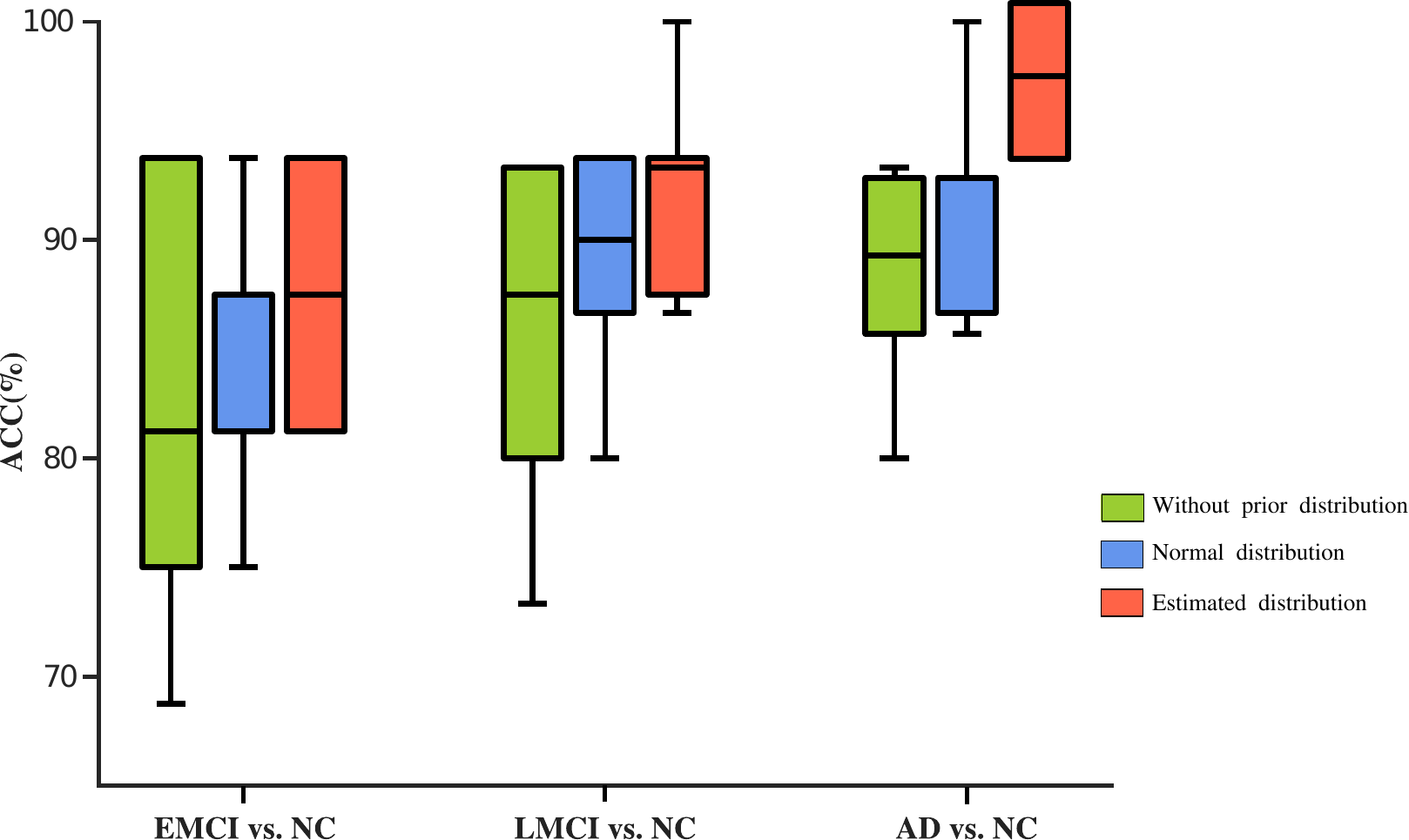}}
	\caption{Effects of prior information on the prediction accuracy.}
	\label{fig3}
\end{figure}

\subsection{Effect analysis of prior distribution}
The prior information is an essential factor to help diagnose disease. An advantage of the proposed network is integrating estimated prior distribution and adversarial network for representation learning. Considering the structure of our network, we investigate the effect of three conditions on final prediction accuracy: 1) without prior distribution, in this situation, the adversarial network is removed from the proposed model; 2) normal prior distribution; 3) estimated prior distribution. The results in our three prediction tasks are shown in Fig.~\ref{fig3}. It can be observed that the network with prior distribution behaves better than the network without prior distribution. Besides, estimating the prior distribution from anatomical knowledge has a more considerable accuracy than the model's performance using normal prior distribution.

\begin{figure}[htbp]
	\centerline{\includegraphics[width=\columnwidth]{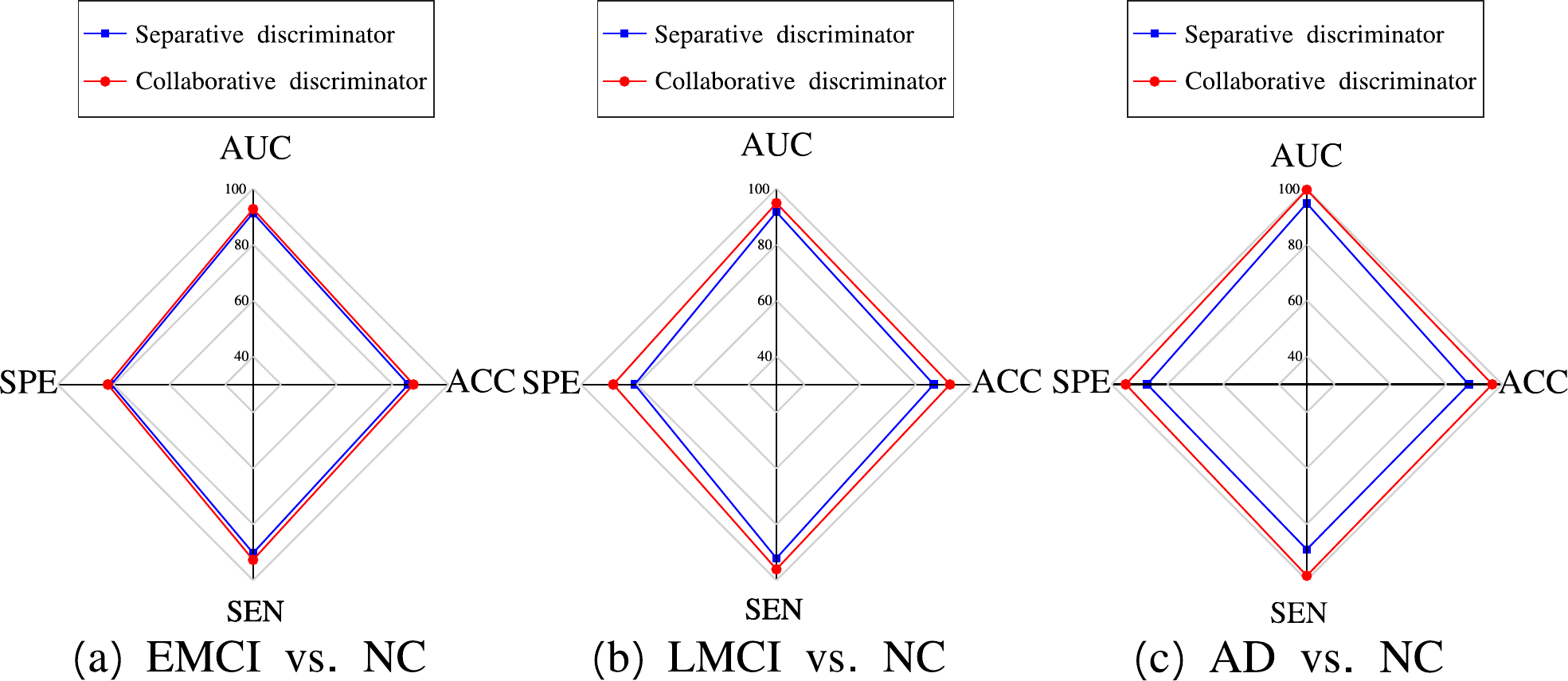}}
	\caption{Effects of the discriminator network on the prediction performance in our three prediction tasks.}
	\label{fig3.5}
\end{figure}

\subsection{Effect analysis of discriminator structure}
The discriminator is significant for the whole adversarial training process. In order to prove the advantages of our designed network in discriminator, we separate it into two individual discriminators for comparison. In this situation, there is no concatenation and summation in Fig.~\ref{fig2}(a).  After the training stage, the same test set of three modal images are used. As is shown in Fig.~\ref{fig3.5}, it can be seen that adding collaborative network in the discriminator achieves better classification performance in terms of AUC, ACC, SPE, and SEN. Overall, the proposed discriminator network can add the joint distribution variability of input features and learned representations to the adversarial optimization process, which in turn brings better performance.

\begin{figure}[htbp]
	\centerline{\includegraphics[width=\columnwidth]{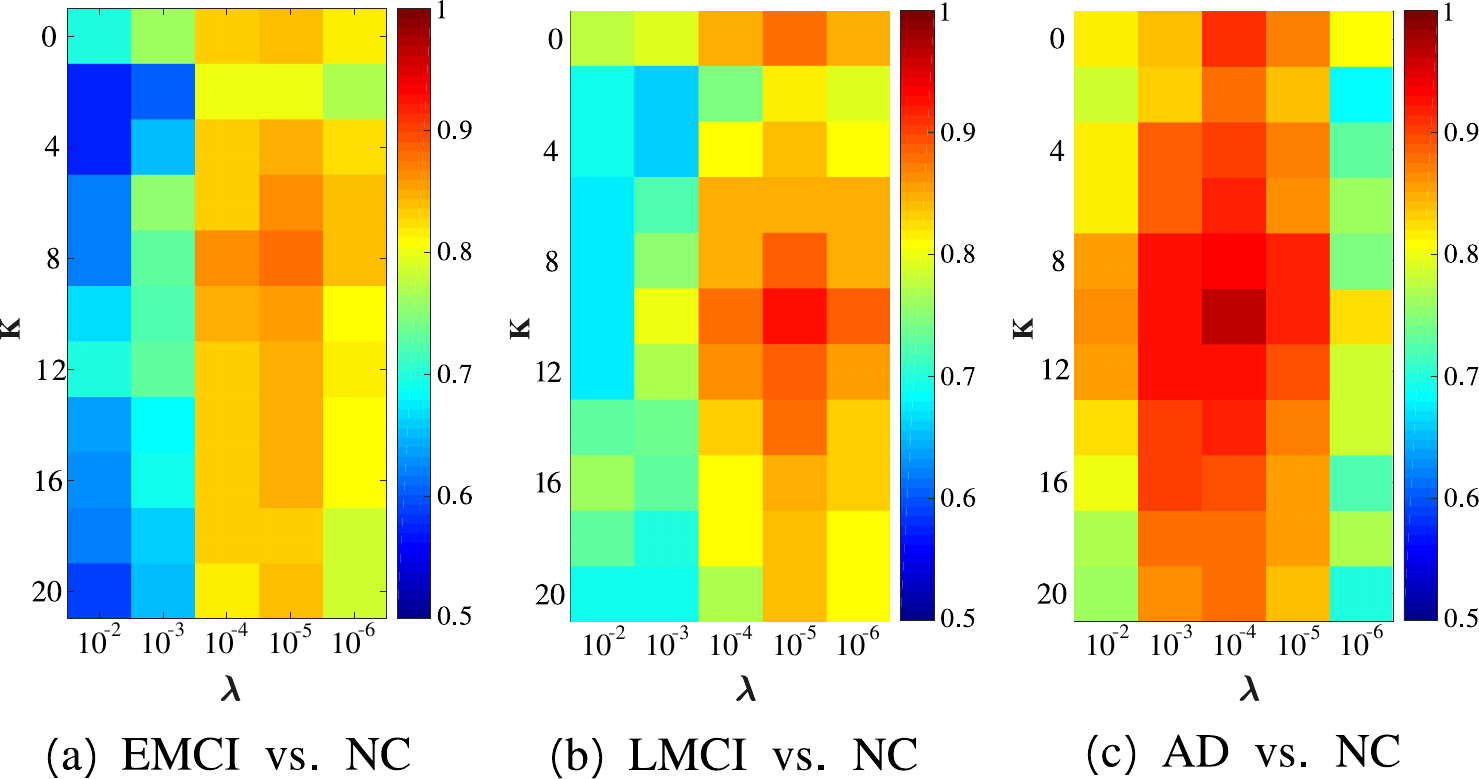}}
	\caption{Effects of hypergraph parameters on the prediction accuracy.}
	\label{fig4}
\end{figure}

\subsection{Effect analysis of hypergraph parameters}

In this section, the effects of hypergraph parameters on the prediction results are investigated to evaluate the performance of the proposed HPN module. The parameter $K$ controls the number of interacting vertices in hypergraph construction, and the parameter $\lambda$ determines the sparsity of the brain united connectivity. As these two parameters both have a great impact on the final results, we evaluated the prediction performance by varying the values of $K$ and $\lambda$ within their respective ranges, i.e., $K\in\{0,2,\cdots,20\}$ and $\lambda\in\{10^{-2},10^{-3},\cdots,10^{-6}\}$. Particularly, the value 0 of $K$ means that the HPN module degrades to graph fusion with normal two-layer GCN. Fig.~\ref{fig4} shows the accuracy of the proposed HPN concerning different combinations of values ($K,\lambda$). The hypergraph method with optimal parameters performs better than the graph method in three classification tasks. The optimal pair of hypergraph parameters are (8,$10^{-5}$) for EMCI vs. NC, (10,$10^{-5}$) for LMCI vs. NC, and (10,$10^{-4}$) for AD vs. NC.

\begin{figure}[htbp]
	\centerline{\includegraphics[width=\columnwidth]{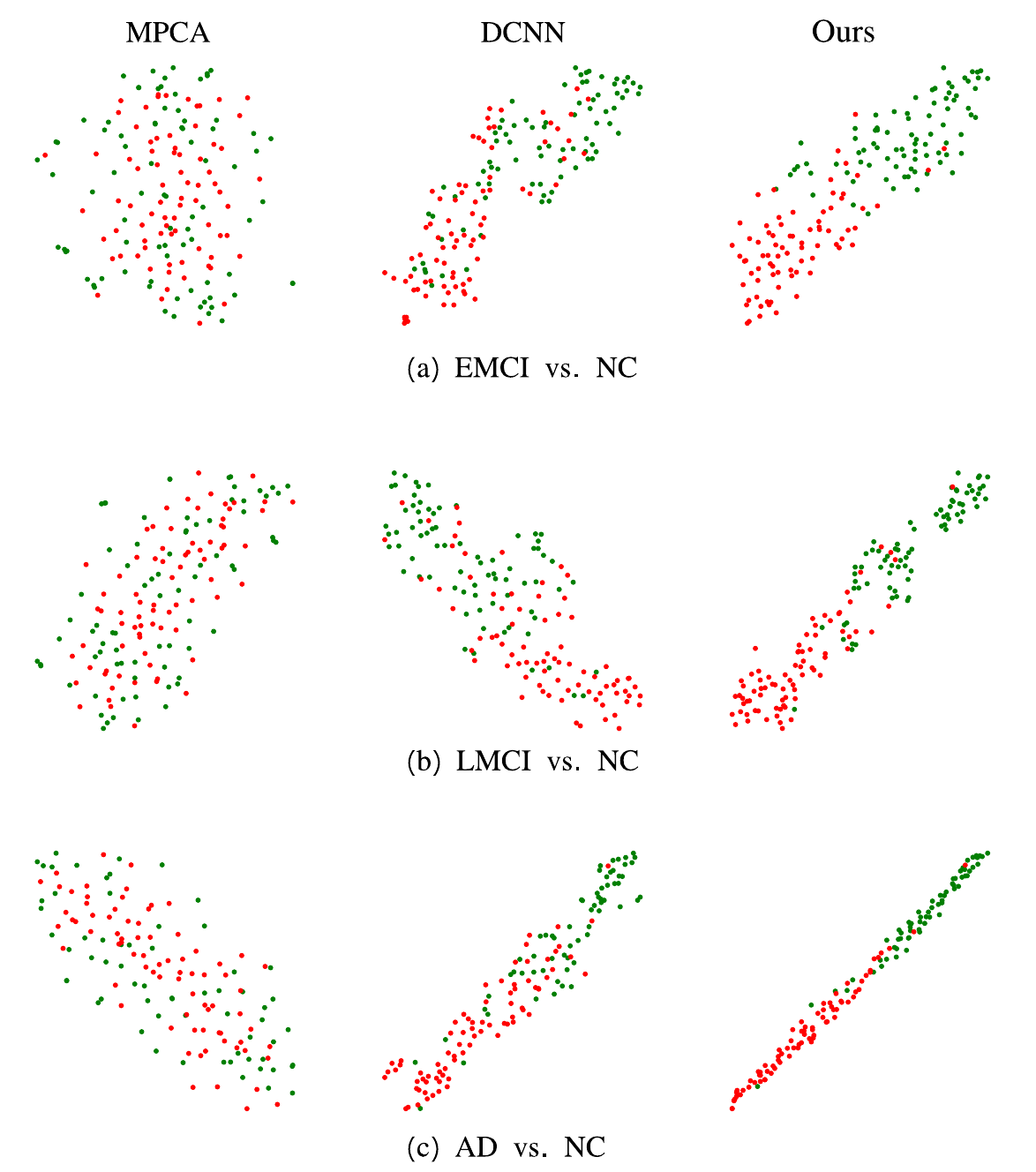}}
	\caption{Visualization of learned features using MPCA, DCNN and Ours for three prediction tasks.}
	\label{fig5}
\end{figure}

\subsection{Prediction results}
The experimental prediction results of all models are displayed in Table~\ref{TableII}. It can be seen that the proposed model using fMRI\&DTI\&MRI achieves the highest mean accuracy of 96.47\%, 92.20\%, and 87.50\% on the prediction of AD vs. NC, LMCI vs. NC, EMCI vs. NC, respectively. It can be concluded that our model performs better than other methods in terms of unimodality, bi-modality or triple-modality medical images. In addition, the results of triple-modality imaging-based methods are superior to that of bi-modality imaging-based methods. In order to analyze the prediction results, fMRI\&DTI based methods are selected to compare the learned latent features. Fig.~\ref{fig5} shows the projection of learned latent features on a two-dimensional plane using t-SNE tools~\cite{ref_49_2}. As can be seen in this figure, the features obtained by our model give the best discriminative map among the three methods.

\begin{figure}[htbp]
	\centerline{\includegraphics[width=\columnwidth]{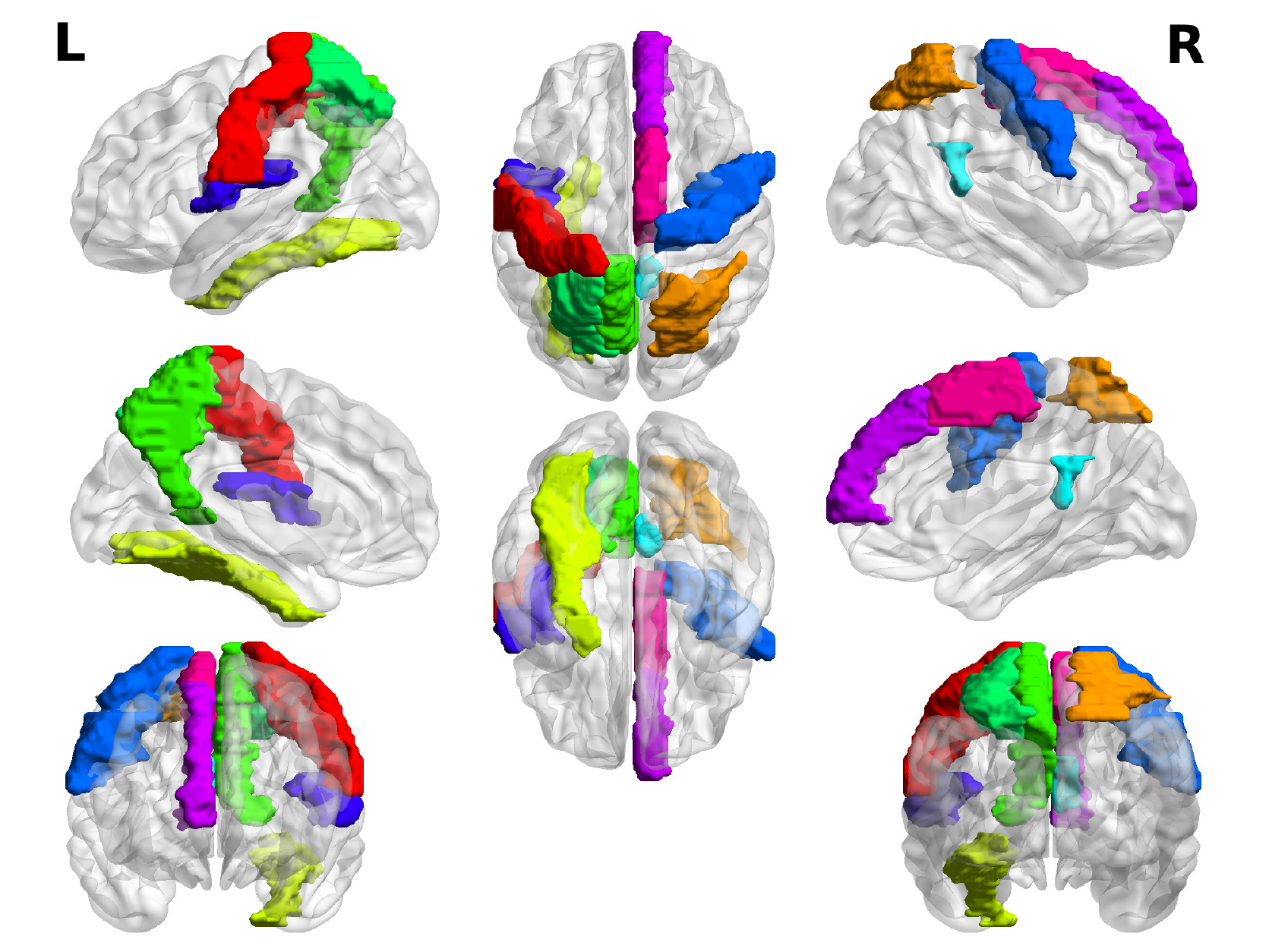}}
	\caption{Visualization of top 10 importance ROIs for EMCI vs. NC.}
	\label{fig6.1}
\end{figure}

\begin{figure}[htbp]
	\centerline{\includegraphics[width=\columnwidth]{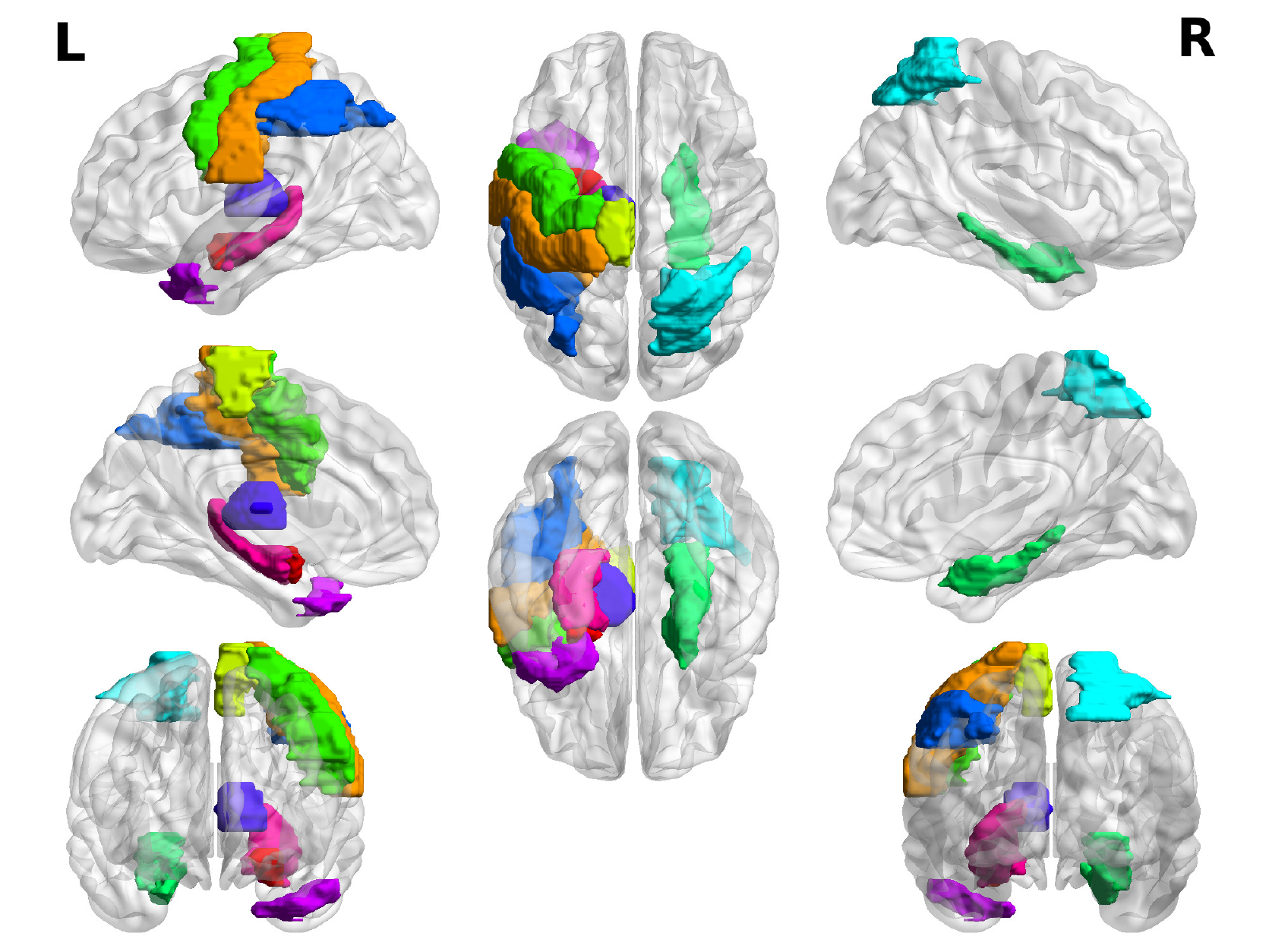}}
	\caption{Visualization of top 10 importance ROIs for LMCI vs. NC.}
	\label{fig6.2}
\end{figure}

\begin{figure}[htbp]
	\centerline{\includegraphics[width=\columnwidth]{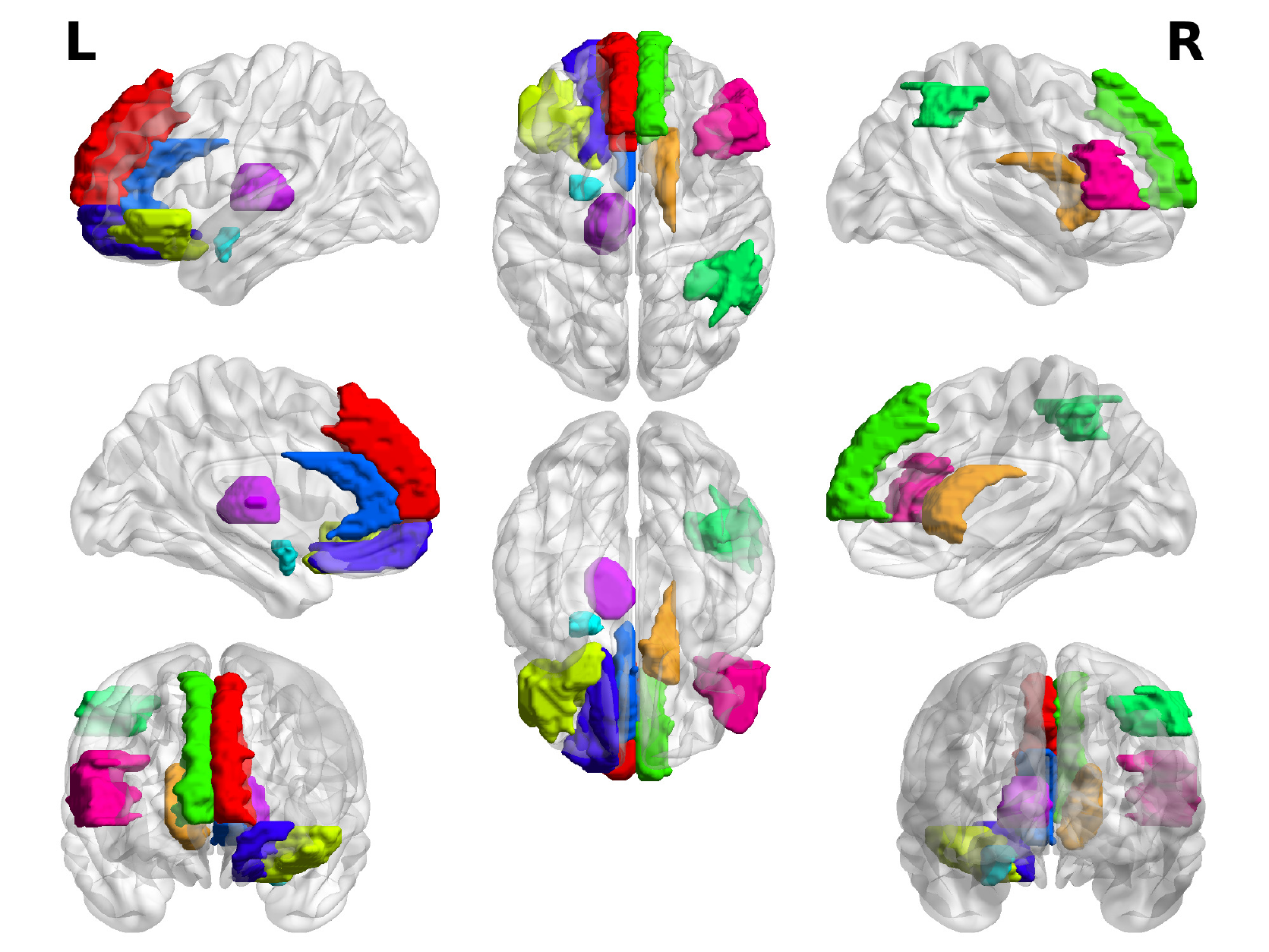}}
	\caption{Visualization of top 10 importance ROIs for AD vs. NC.}
	\label{fig6.3}
\end{figure}

\begin{figure}[htbp]
	\centerline{\includegraphics[width=\columnwidth]{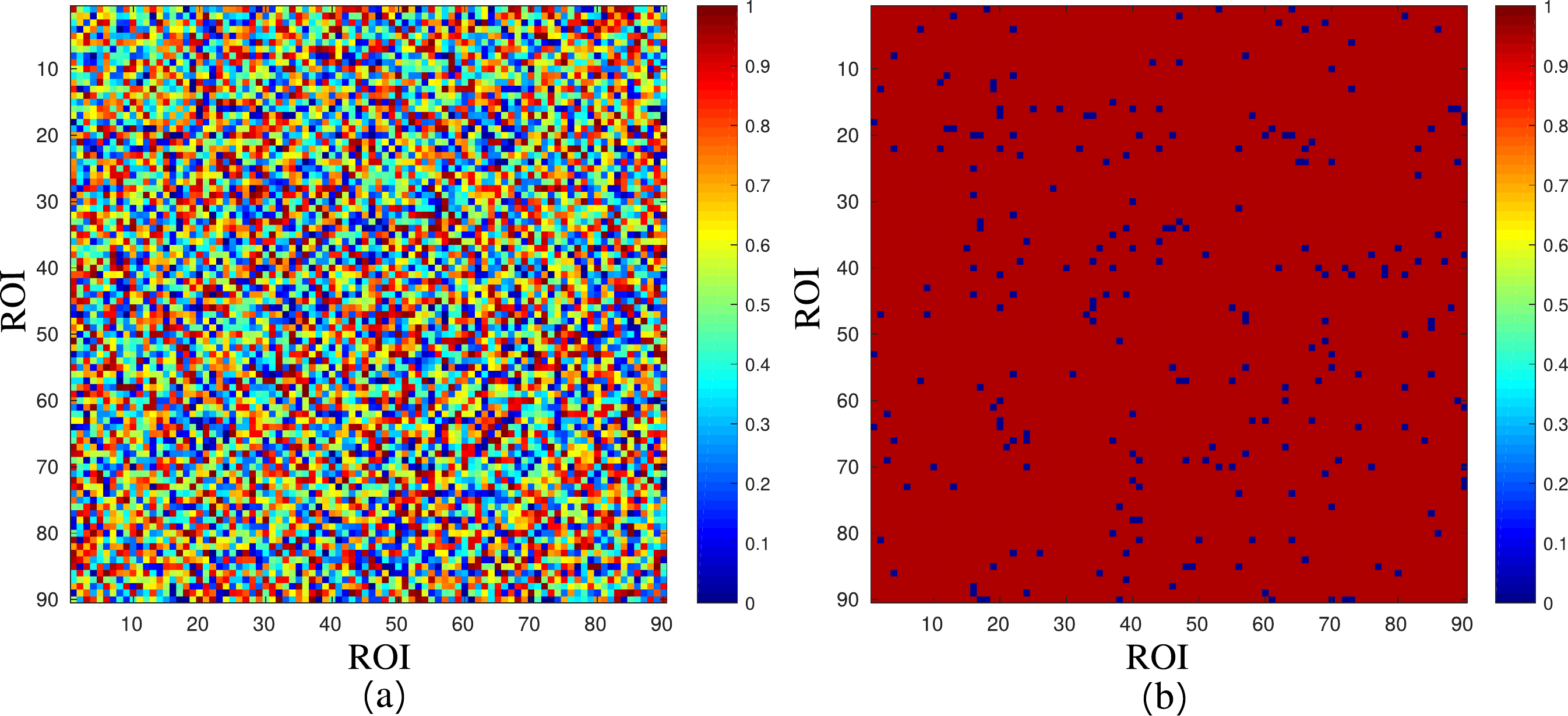}}
	\caption{Visualization of $p$-values on connections between ROIs for EMCI vs. NC using the proposed model. (a) The $p$-values on connections between ROIs. (b) The connections with $p$-value less than 0.05.}
	\label{fig8.1}
\end{figure}

\begin{figure}[htbp]
	\centerline{\includegraphics[width=\columnwidth]{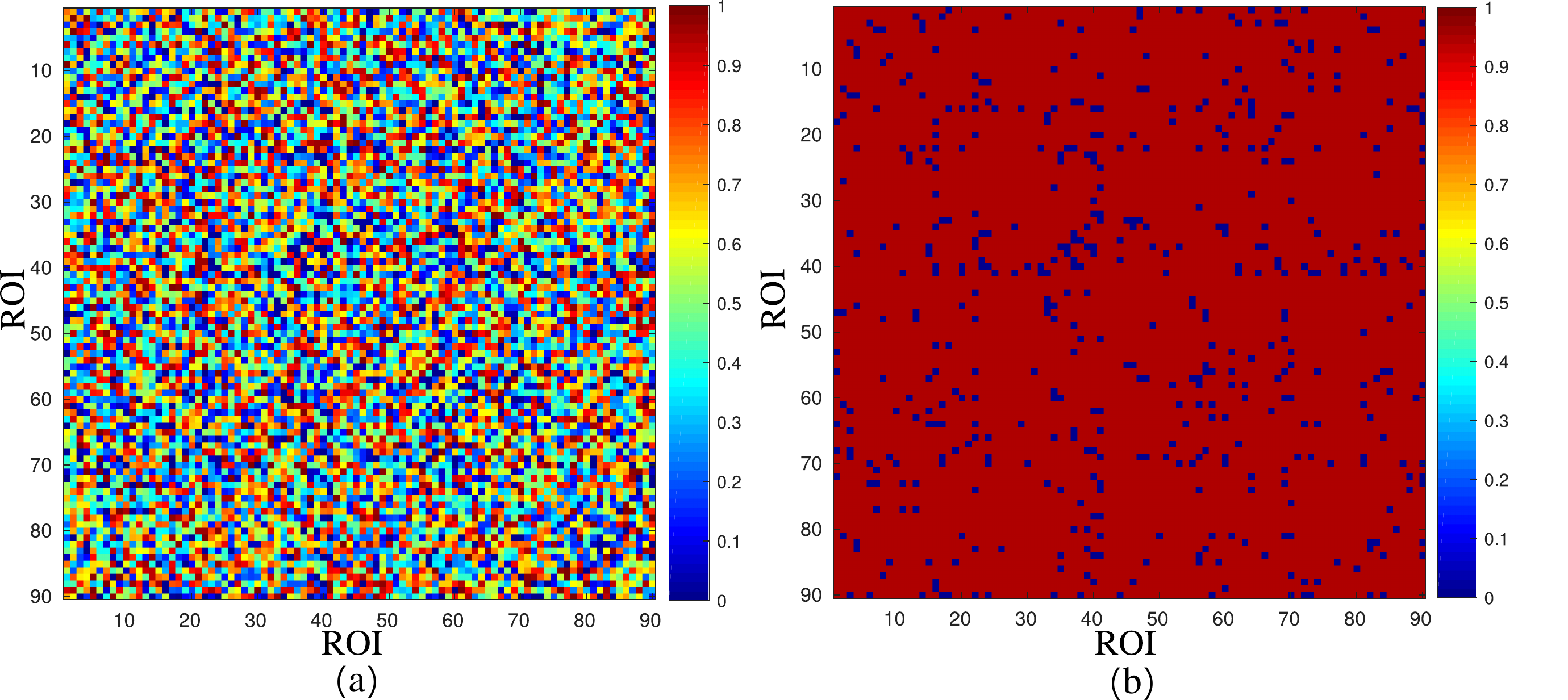}}
	\caption{Visualization of $p$-values on connections between ROIs for LMCI vs. NC using the proposed model. (a) The $p$-values on connections between ROIs. (b) The connections with $p$-value less than 0.05.}
	\label{fig8.2}
\end{figure}

\subsection{Quantitative analysis of important brain regions}
To evaluate the influence of different brain regions on the prediction tasks, we shield one brain region and get one importance score for that brain region by using a 10-fold cross-validation strategy. The importance score is defined as one minus the mean accuracy. Hence, a considerable value of importance score means a critical brain region. According to the 90 ROIs in the AAL atlas, we sort the importance scores and obtain ten important brain regions for each prediction task. Fig.~\ref{fig6.1}, Fig.~\ref{fig6.2} and Fig.~\ref{fig6.3} shows the top 10 ROIs in terms of their importance scores for EMCI vs. NC, LMCI vs. NC and AD vs. NC, respectively. The 10 important brain regions of EMCI vs. NC are PoCG.L, SPG.R, FFG.L, PCUN.L, SPG.L, PCG.R, ROL.L, SFGmed.R, SMA.R and ITG.L. The ten important brain regions of LMCI vs. NC are AMYG.L, PoCG.L, PCL.L, PreCG.L, PHG.R, SPG.R, IPL.L, THA.L, TPOmid.L, and HIP.L. The 10 important brain regions of AD vs. NC classification are SFGmed.L, CAU.R, ORBinf.L, SFGmed.R, IPL.R, AMYG.L, ACG.L, ORBsup.L, THA.L and IFGtriang.R.

\begin{figure}[htbp]
	\centerline{\includegraphics[width=\columnwidth]{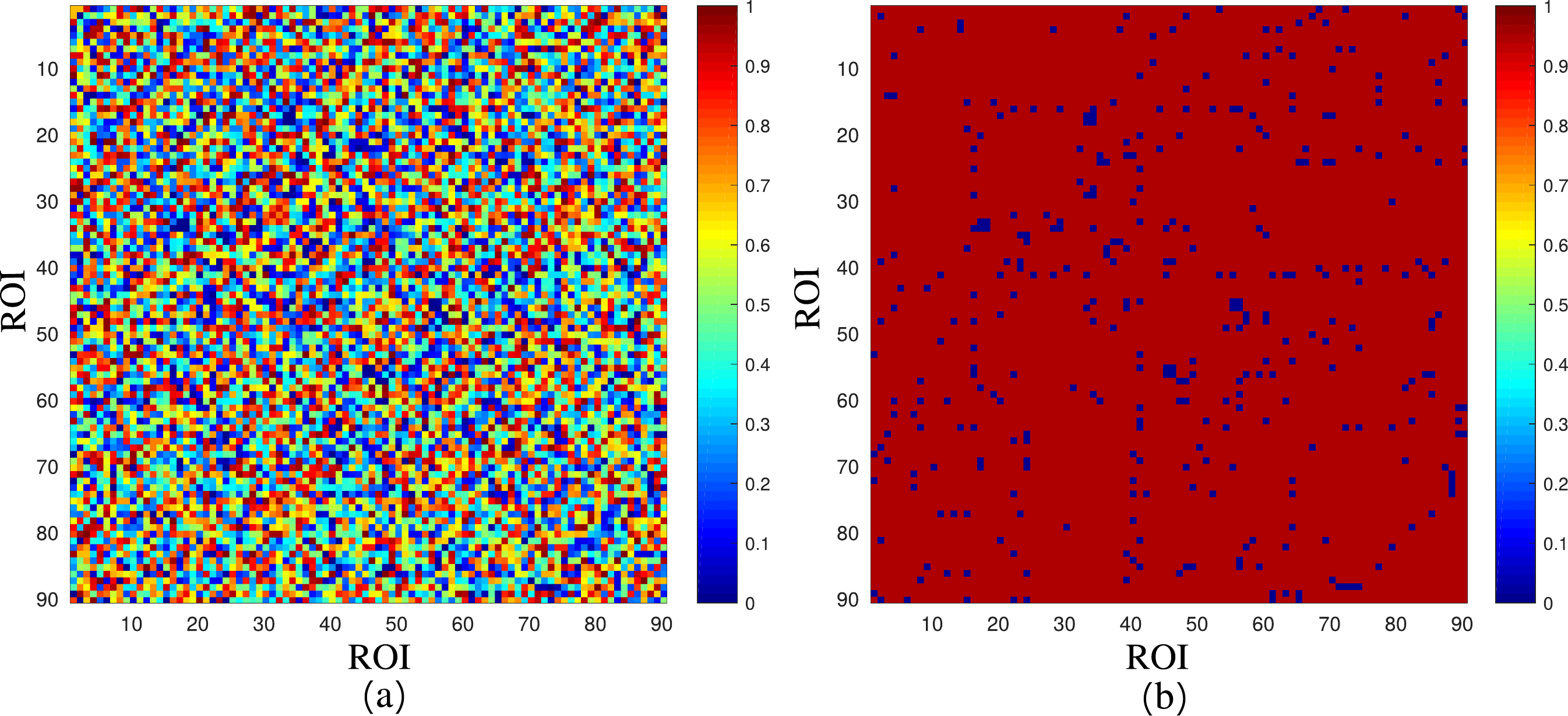}}
	\caption{Visualization of $p$-values on connections between ROIs for AD vs. NC using the proposed model. (a) The $p$-values on connections between ROIs. (b) The connections with $p$-value less than 0.05.}
	\label{fig8.3}
\end{figure}

\begin{figure}[htbp]
	\centerline{\includegraphics[width=\columnwidth]{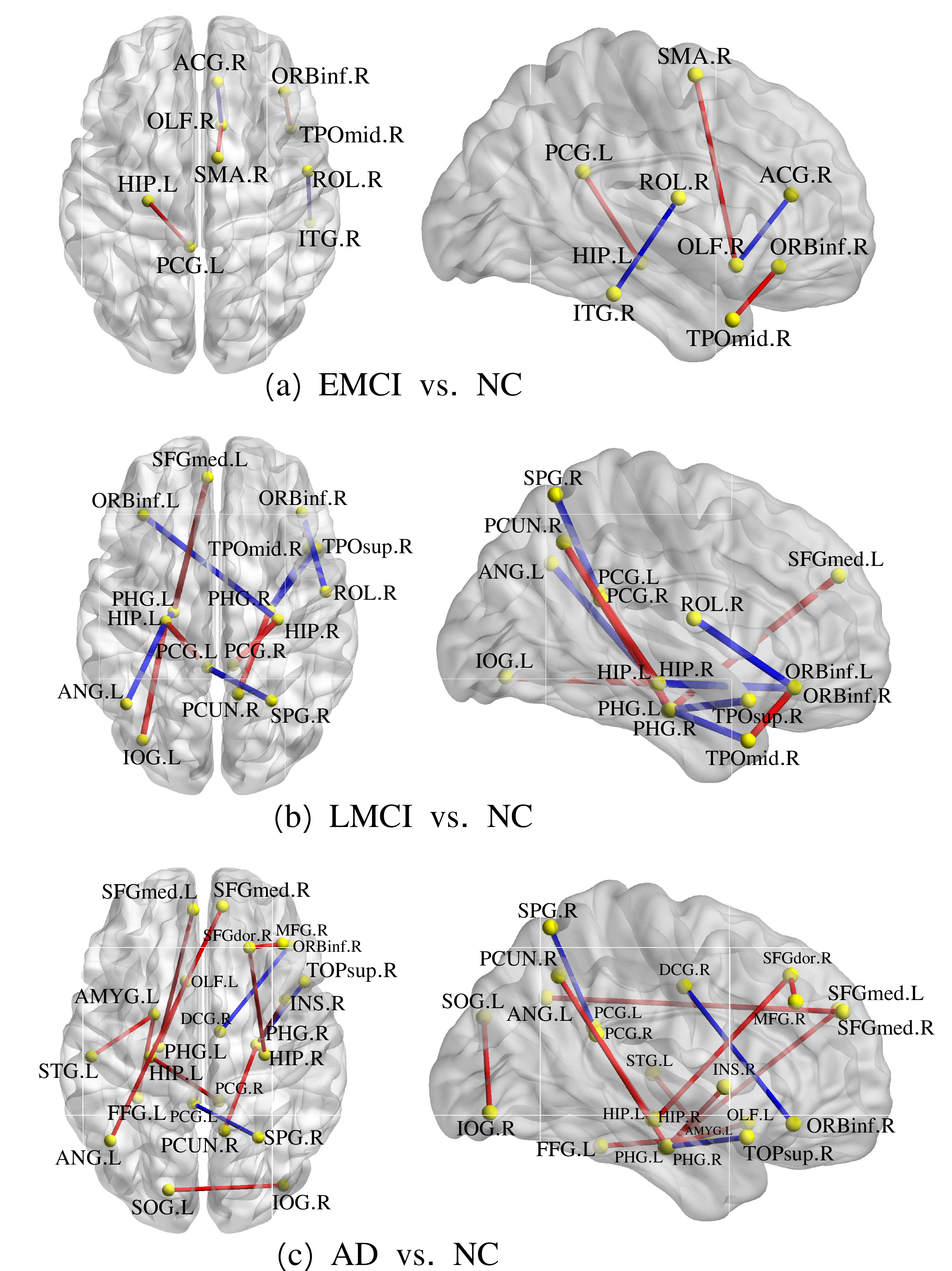}}
	\caption{Visualization of important abnormal connections between two groups with $p$-value less than 0.001. }
	\label{fig9}
\end{figure}

\begin{figure*}[htbp]
	\centering
	\includegraphics[scale=0.58]{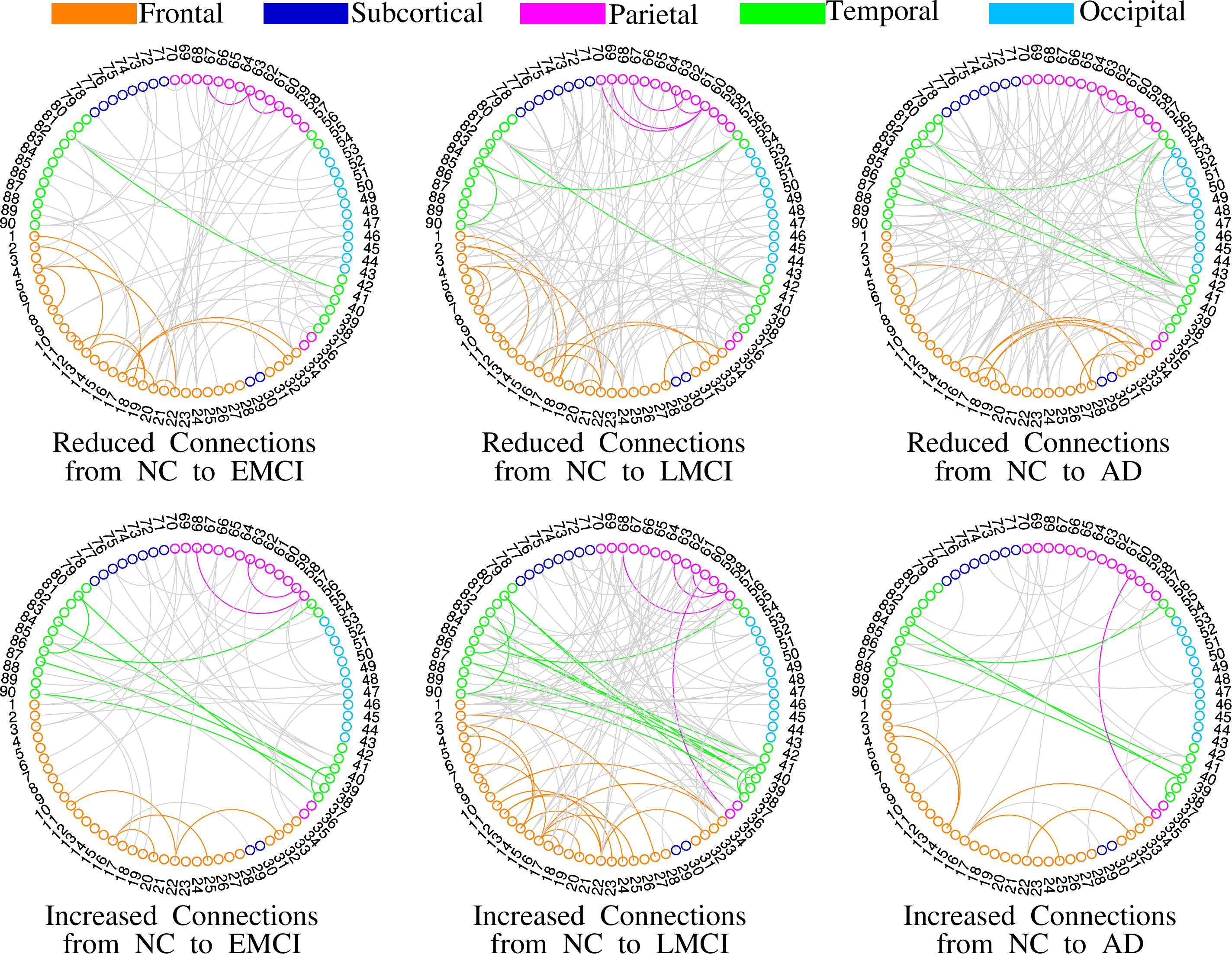}
	\caption{Visualization of altered connections for EMCI vs. NC, LMCI vs. NC, and AD vs. NC. Connections between intra-network are marked in color, while grey means connections between inter-network. Note that, the 90 ROIs are spitted into 5 brain regions. The intra-network means connections within the same brain region, and the inter-network means connections between different brain regions.}
	\label{fig10}
\end{figure*}

\subsection{Quantitative analysis of abnormal brain connections}
To uncover the mechanism of cognitive disease, we investigate the obtained united connectivity (UC) using statistical tests between two different groups. In particular, we constructed a UC for each subject using fMRI\&DTI\&MRI and then estimated the significance of each connection between two groups using the standard two-sample t-test method. Fig.~\ref{fig8.1}, Fig.~\ref{fig8.2} and Fig.~\ref{fig8.3} show the $p$-values of connections between each pair of ROIs for EMCI vs. NC, LMCI vs. NC and AD vs. NC, respectively. For ease of visualization, a threshold $p$-value with 0.05 is chosen to display the significant connections. For each prediction task, we count the significant connections for each ROI and then obtain 10 top discriminative ROIs with the highest occurrence frequency. The significant connections and ROIs of EMCI vs. NC are mainly located in SMA.R, OLF.R, ORBinf.R, PHG.R, ITG.R, PCL.R, ROL.L, SFGmed.R, PoCG.L, ANG.R. The significant brain regions for the task of LMCI vs. NC are OLF.R, AMYG.L, ITG.R, ORBinf.R, PHG.R, PCL.L, HIP.L, PCL.R, FFG.R, and DCG.R. The significant brain regions for the task of AD vs. NC are AMYG.L, ORBinf.R, FFG.R, DCG.R, SPG.R, OLF.R, SFGmed.R, PHG.L, LING.R, and SMG.R. The list regions are partly overlapped with the results in the above section, and these findings are partly consistent with previous studies~\cite{ref_30,ref_39}. In order to visualize the important abnormal connections at different stages, we choose $p$-values smaller than 0.001 to select significant connections between two groups. Those important abnormal connections are displayed in Fig.~\ref{fig9} using the BrainNet Viewer~\cite{ref_50}.

\begin{figure}[htbp]
	\centerline{\includegraphics[width=\columnwidth]{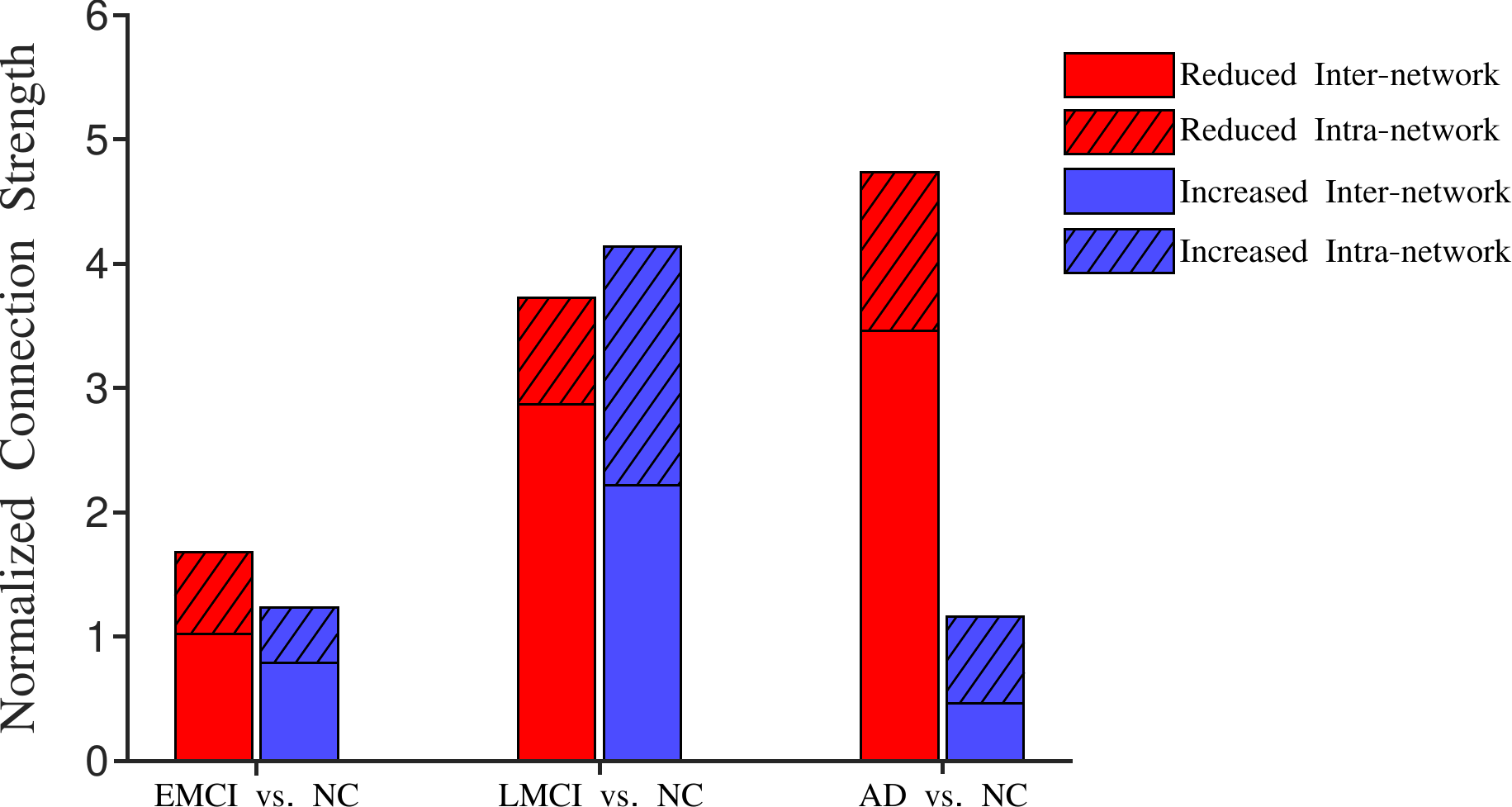}}
	\caption{Normalized altered connection strength in three stages of the cognitive disease with respect to NC.}
	\label{fig11}
\end{figure}

Based on the significant $p$-values, we calculate the mean connection strength for each group to analyze the characteristics of subjects at different stages. We mean the UCs for each group and then subtract the mean UC of the NC group from the mean UC of the patient's group (i.e., EMCI, LMCI, AD). The subsequent connectivity means altered connections. As is displayed in Fig.~\ref{fig10}, the altered connections differ in different stages. Specifically, the number of reduced connections rises from EMCI to AD stage, while the number of increased connections rises from EMCI stage to LMCI stage and then drops down at AD stage. In addition, most of the altered connections are between inter-networks, regardless of the connection number or the connection strength. Fig.~\ref{fig11} displayed the normalized connection strength of inter-networks and intra-networks in different prediction tasks. The reduced connection strength of the AD stage is the largest in the three stages, while the most considerable increased connection strength emerges at the LMCI stage.

\section{Discussion}
In the work of abnormal connection prediction, the characteristics change in connections strength suggest a compensate mechanism~\cite{ref_51,ref_52} at the MCI stage and a deteriorate progression at the AD stage. As shown in Fig.~\ref{fig11}, at the early stage of AD, some brain connections are damaged while some other connections emerged or are strengthened to compensate for functional activities. With the progression of the disease, more extensive and severe network damage occurs at the late stage of AD, which leads to the degeneration of the whole brain function and, finally, severe cognitive decline. Most of the reduced connections occur in inter-network, and the increased connections are likely to emerge in intra-network. Furthermore, part of our discovered abnormal connections at the MCI stage is consistent with the neuroscience findings~\cite{ref_54}. Relative to NC, the left hippocampus at the LMCI stage lose connections to the left parahippocampal gyrus and left angular gyrus while making new connections to the left posterior cingulate gyrus and left inferior occipital gyrus. Other reduced connections could be identified between the right parahippocampal gyrus and the right temporal pole: superior temporal gyrus, as well as the right temporal pole: middle temporal gyrus. Also, the rest identified increased connections are between the right hippocampus and the right posterior cingulate gyrus, and additionally between the left parahippocampal gyrus and left medial superior frontal gyrus. The Fig.~\ref{fig9} also suggests that reduced connections tend to have longer distances than increased connections, which is consistent with the work in ~\cite{ref_55,ref_56}.

%\subsection{Limitations and Future Research}
There are still two limitations in the current study. One is that we introduce only four known closely diseased-related ROIs to estimate the prior distribution. Some diseased-unrelated ROIs will be considered in future work. Another one is that the data used in this study are mainly neuroimaging data. Since genetic modality is correlated with cognitive disease~\cite{ref_57}, it is interesting to add single nucleotide polymorphism (SNP) data for abnormal connection prediction in future studies.

\section{Conclusion}
In this paper, we proposed a PGARL-HPN for predicting abnormal brain connections at different stages of AD, which effectively integrates fMRI, DTI, and MRI. As an attempt to make use of prior distribution estimated from anatomical knowledge, a bidirectional adversarial mechanism with a novel pairwise collaborative discriminator was designed to help the generator learn joint representations while keeping the learned representations in the same distribution. Also, the HPN module was developed to effectively fuse the learned representations and capture the complementary multimodal information. Quantitatively experimental results suggest that our proposed model performs better than other related methods in analyzing and predicting Alzheimer's disease progression. Moreover, the proposed model can evaluate characteristics of abnormal brain connections at different stages of Alzheimer's disease, where part of the identified abnormal connections are consistent with previous neuroscience discoveries. The identified abnormal connections may help us understand the underlying mechanisms of neurodegenerative diseases and provide biomarkers for early cognitive disease treatment.

\section*{Acknowledgment}
This work was supported by the National Natural Science Foundations of China under Grant 61872351 and the International Science and Technology Cooperation Projects of Guangdong under Grant 2019A050510030.

\end{document}